\documentclass[letterpaper, 10 pt, conference]{ieeeconf}  

\usepackage{etoolbox}
\newtoggle{ext}     
\toggletrue{ext}    

\IEEEoverridecommandlockouts                              

\usepackage{graphics} 
\usepackage{epsfig} 
\usepackage{times} 
\usepackage{amsmath,amssymb}
\DeclareMathAlphabet{\pazocal}{OMS}{zplm}{m}{n}
\usepackage[style=ieee,url=false,doi=false]{biblatex}
\usepackage[export]{adjustbox}
\usepackage{booktabs}
\usepackage{hyperref}
\usepackage{cleveref}
\usepackage{setspace}
\usepackage{tabularx}
\usepackage{enumerate}
\usepackage{newtxtext,newtxmath}

\iftoggle{ext}{}{
\setlength{\parindent}{1em}
\usepackage[compact]{titlesec}
\titlespacing{\subsection}{0pt}{1ex}{0ex}

\renewcommand{\baselinestretch}{0.99}
\setlength{\abovedisplayskip}{0pt} \setlength{\abovedisplayshortskip}{0pt}
\setlength{\belowdisplayskip}{0pt} \setlength{\belowdisplayshortskip}{0pt}
}

\usepackage{mathtools}
\DeclareMathAlphabet\mathbfcal{OMS}{cmsy}{b}{n}

\newtheorem{defn}{Definition}

\providecommand{\R}{\ensuremath \mathbb{R}}

\providecommand{\N}{\ensuremath \mathbb{N}}

\newcommand{\ie}{\textit{i.e., }}
\newcommand{\eg}{\textit{e.g., }}

\newcommand{\norm}[1]{\left\Vert#1\right\Vert}

\newcommand{\defeq}{\vcentcolon=}

\DeclareMathOperator*{\minimize}{\mathrm{minimize}}
\DeclareMathOperator*{\subjectto}{\mathrm{subject~to}}
\DeclareMathOperator*{\argmin}{arg\,min}

\newcommand{\fstate}{\mathbf{x}^{f}}
\newcommand{\rstate}{\mathbf{x}}
\newcommand{\perf}{\mathbf{z}}
\newcommand{\obs}{\mathbf{y}}
\newcommand{\perfref}{\bar{\perf}}

\newcommand{\autSSM}{\pazocal{W}(E)}

\newcommand{\mass}{\mathbf{M}}
\newcommand{\damp}{\mathbf{C}}
\newcommand{\stiff}{\mathbf{K}}
\newcommand{\q}{\mathbf{q}}
\newcommand{\ctrl}{\mathbf{u}}
\newcommand{\Cntrl}{\pazocal{U}}
\newcommand{\Observable}{\pazocal{Z}}
\newcommand{\Q}{\mathbf{Q}}
\newcommand{\Qf}{\mathbf{Q}_{\mathrm{f}}}
\newcommand{\Rcost}{\mathbf{R}}
\newcommand{\spectralsubspace}{E}
\newcommand{\ytox}{\mathbf{v}}
\newcommand{\xtoy}{\mathbf{w}}

\newcommand{\Ydata}{\mathbfcal{Y}}
\newcommand{\X}{\mathbfcal{X}}
\newcommand{\U}{\mathbf{U}}
\newcommand{\Br}{\mathbf{B}_r}

\newcommand{\nfstate}{{n_{f}}}
\newcommand{\nrstate}{n}
\newcommand{\nctrl}{m}
\newcommand{\nperf}{o}
\newcommand{\nout}{p}

\newcommand{\nodes}{\pazocal{N}}

\title{\LARGE \bf
Data-Driven Spectral Submanifold Reduction for \\ Nonlinear Optimal Control of High-Dimensional Robots
}

\author{John Irvin Alora$^{1}$, Mattia Cenedese$^{2}$, Edward Schmerling$^{1}$, George Haller$^{2}$, Marco Pavone$^{1}$%
\thanks{J.A. is supported by the Secretary of the Air Force STEM Ph.D. Fellowship. This work was supported by the NASA University Leadership Initiative (grant \#80NSSC20M0163) and KACST; this article solely reflects the opinions and conclusions of its authors and not any Air Force, NASA, nor KACST entity.}
\thanks{
$^{1}$Department of Aeronautics and Astronautics, Stanford University, Stanford, CA, 94305, USA {\tt\small\{jjalora, schmrlng, pavone\}@stanford.edu}}
\thanks{
$^{2}$Institute for Mechanical Systems, ETH Zurich, 8092 Z\"urich, Switzerland {\tt\small\{mattiac, georgehaller\}@ethz.ch}}
}

\begin{document}

\makeatletter
\patchcmd{\@makecaption}
  {\scshape}
  {}
  {}
  {}
\makeatletter
\patchcmd{\@makecaption}
  {\\}
  {.\ }
  {}
  {}
\makeatother
\def\tablename{Table}

\maketitle
\thispagestyle{empty}
\pagestyle{empty}

\begin{abstract}

Modeling and control of high-dimensional, nonlinear robotic systems remains a challenging task.
While various model- and learning-based approaches have been proposed to address these challenges, they broadly lack generalizability to different control tasks and rarely preserve the structure of the dynamics.
In this work, we propose a new, data-driven approach for extracting low-dimensional models from data using Spectral Submanifold Reduction (SSMR).
In contrast to other data-driven methods which fit dynamical models to training trajectories, we identify the dynamics on generic, low-dimensional attractors embedded in the full phase space of the robotic system.
This allows us to obtain computationally-tractable models for control which preserve the system's dominant dynamics and better track trajectories radically different from the training data.
We demonstrate the superior performance and generalizability of SSMR in dynamic trajectory tracking tasks \textit{vis-\'a-vis} the state of the art, including Koopman operator-based approaches.

\end{abstract}

\section{INTRODUCTION}
High-dimensional robotic systems promise to revolutionize the field of robotics due to the versatility brought forth by their large degrees of freedom (DOF).
For example, continuum soft robots can exhibit \textit{embodied intelligence} \cite{mengaldo2022concise} in which they conform to surfaces and objects while maintaining a level of physical robustness unavailable to their more rigid counterparts.
This level of compliance and elasticity make them well-suited to operate in delicate, geometrically constrained environments, which enable them to play crucial roles in settings where safe human-robot interaction is paramount. 

Unfortunately, these advantages pose significant practical challenges for the modeling and control of these robots. This is due to the inherent nonlinearities and high DOF required to accurately capture the structural deformations that realize these compliant behaviors.
While several model- and learning-based approaches have been proposed in literature to address some of these challenges, these methods suffer from their inability to tractably bridge the gap between having accurate, but low-dimensional models.
This accuracy--dimensionality tradeoff results in methods that sacrifice predictive accuracy and structure preservation for a drastic decrease in dimensionality or vice versa.

Motivated by recent developments in \textit{Spectral Submanifold} (SSM) theory \cite{haller2016nonlinear} and its successful application to data-driven predictions of nonlinearizable phenomena \cite{cenedese2022ssmlearn}, we propose a new data-driven \textit{Spectral Submanifold Reduction} (SSMR) framework for learning low-dimensional, faithful dynamics of high-dimensional robots on SSMs.
SSMs, as summarized in Figure~\ref{fig:SSM}, are low-dimensional, attracting invariant manifolds which capture highly nonlinear phenomena of high-dimensional systems.
By learning the dynamics on these generic structures, we extract low-dimensional, control-oriented models that preserve the dominant physics of the system.
This allows SSMR to overcome common drawbacks associated with data-driven approaches such as lack of generalizability, high-data requirement, and sensitivity to noise.


\textbf{Statement of Contributions}:
\begin{figure}
    \centering
    \includegraphics[width=0.88\columnwidth]{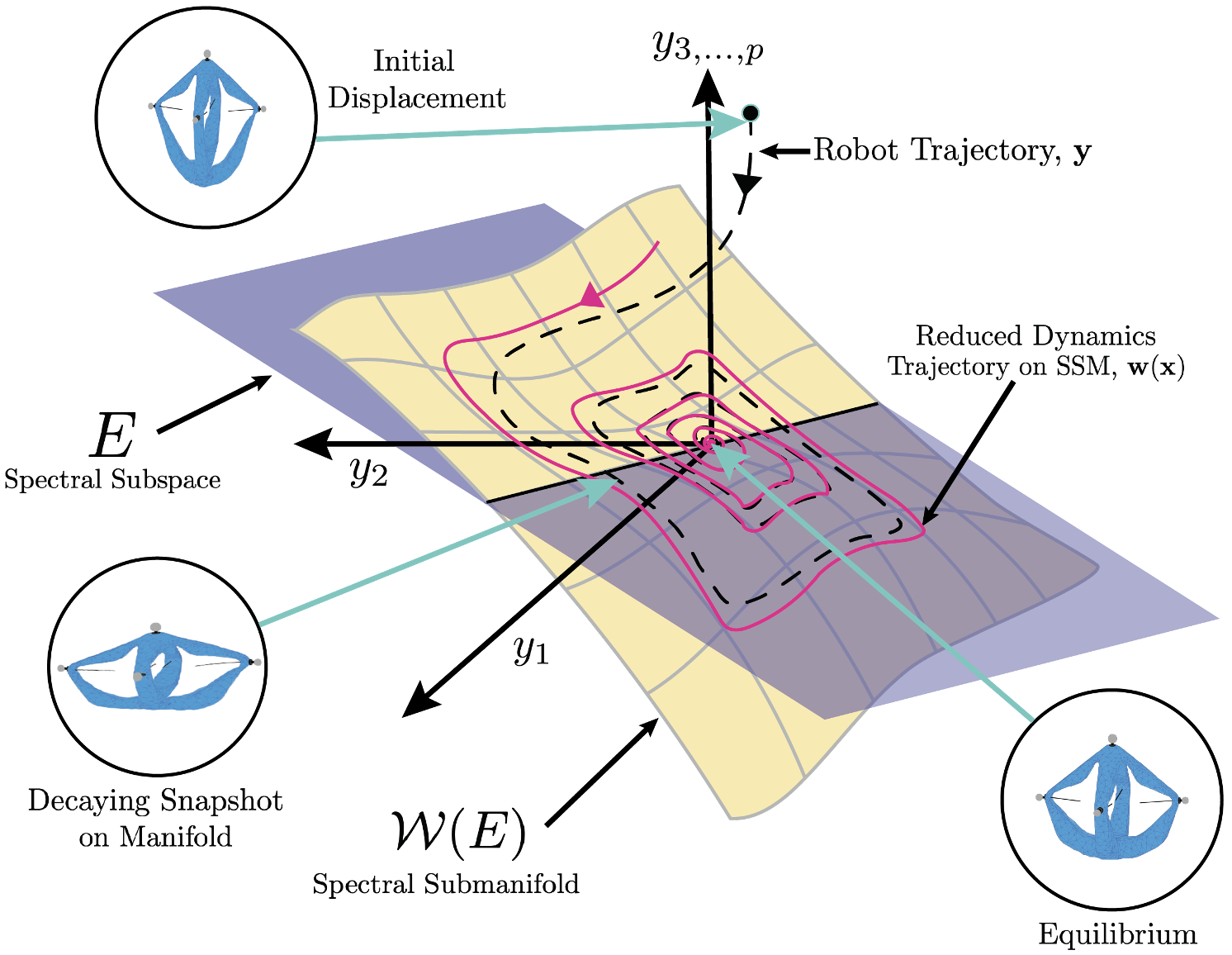}
    \caption{The SSM is a low-dimensional invariant manifold in the robot's phase space which exponentially attracts full state trajectories, causing them to synchronize with the persistent dynamics on the SSM. These structures can capture highly nonlinear behaviors far away from the fixed point and can be approximated arbitrarily-well without increasing the dimension of the SSM.}
    \label{fig:SSM}
    \vspace{-0.6cm}
\end{figure}
(i) We present SSMR, the first data-driven approach for learning the dynamics of high-dimensional robots on SSMs for control. We extend recent work on \textit{SSMLearn} \cite{cenedese2022data} by providing the additional innovation of disambiguating the effect of control from the underlying dynamics on the SSM. This allows us to extract highly-accurate models for control in an \textit{equation-free} manner.

(ii) We extend previous work on SSM-based control \cite{mahlknecht2022ssm} to general control tasks by implementing a SSMR optimal control scheme and validating it on simulations of a high-dimensional soft robot. We show that SSMR outperforms the state-of-the-art methods in both trajectory tracking performance and computational efficiency, highlighting a key feature of SSMR: it neither compromises accuracy nor computational tractability.

\textbf{Related Work}: While prevailing approaches for modeling high-dimensional robots involve the use of simplified assumptions that approximate the robot’s behavior \cite{chirikjian1994hyper,rucker2010equilibrium,jung2011modeling}, these methods are only accurate for specific types of geometries and their low fidelity precludes their use in more challenging control tasks.
To address this issue, a popular approach is to compress the governing equations of high-fidelity computational models using projection-based model reduction to produce low-dimensional control surrogates.
While this approach has seen some success for the control of linearized systems \cite{thieffry2018control, katzschmann2019dynamically, della2020model}, model validity deteriorates rapidly when far away from the linearization point.
An attempt at capturing the nonlinearities of the high-dimensional system via piecewise-affine approximation was proposed in \cite{TonkensLorenzettiEtAl2021}, while the work in \cite{goury2018fast} uses an energy-conserving mesh sampling and weighing scheme \cite{farhat2015structure} to construct a reduced order model for inverse-kinematic control.
A common limitation of these projection-based techniques is that the accuracy of the low-dimensional surrogate depends on the choice and size of the subspace, which can grow rapidly for only incremental improvement in accuracy.
Additionally, since these approaches require knowledge of the governing equations, their application to real-world robots remain challenging.
Indeed, the process of extracting accurate models from finite element code is an encumbering and code-intrusive process.

\iftoggle{ext}{
Much of the literature in this direction is focused on using neural networks (NN) for learning approximations of these high-dimensional dynamics from observed transitions. 
From black-box architectures using simple multilayer perceptrons \cite{thuruthel2018model} to grey-box architectures that aim to preserve physical invariants \cite{greydanus2019hamiltonian,cranmer2020lagrangian}, these approaches vary by the level of inductive bias they introduce. In many cases the high-dimensionality of the dynamics stems from the fact that the learning problem is posed in terms of high-dimensional observations (\eg pixel images); the assumption is that there exist underlying low-dimensional latent state-space dynamics, to be learned, that explain the observations \cite{ichter2019robot,fries2022lasdi}. This is similar to the setting of this work, where a core assumption is that the principal dynamics of, \eg a continuum soft robot live on an underlying low-dimensional manifold. For such dissipative physical systems, critically, SSM theory yields insights on the manifold structure that we use to design the learning methodology.
}{
To overcome these challenges, there has been increasing interest in using machine learning techniques to construct data-driven models of high-dimensional robots. 
Much of the literature in this direction is focused on using different neural network (NN) architectures with varying levels of inductive bias, for learning approximations of these high-dimensional dynamics from observed transitions \cite{thuruthel2018model, greydanus2019hamiltonian, cranmer2020lagrangian}.
}

Recently, the Koopman operator has attracted significant interest in the robotics community for data-driven learning of nonlinear dynamics. For example, finite-dimensional, data-driven approximations of (infinite-dimensional) Koopman operators were shown to outperform standard NN models for predicting soft robot dynamics in \cite{bruder2019nonlinear}. 
Since observed dynamics under the Koopman operator are linear, the approach lends itself to established control techniques such as model predictive control (MPC), as shown in \cite{bruder2019modeling}. 
Although this approach is conceptually appealing, most physical systems do not admit exact finite-dimensional, linear representations \cite{brunton2016koopman}.  
Similar to the projection-based methods, the Koopman approach suffers from the accuracy--dimensionality tradeoff since, in theory, more accurate Koopman models require increased number of \textit{a-priori} chosen observable functions.

A common drawback with current data-driven approaches is that they typically result in models that rarely preserve all, if any, of the inherent structure of the dynamics (\eg structural modes, passivity, etc). 
This results in models that do not generalize well to control tasks which involve trajectories outside the training set.
Many of these approaches are also data-intensive and sensitive to noise; 
this precludes their in the design process and achieving good control performance in closed-loop requires a significant tuning of various hyperparameters.
By explicitly targeting rigorous and generic structures in the high-dimensional robot's phase space, we are able to extract reduced-order models which overcome these issues.

\textbf{Organization}: We begin in Section \ref{sec:problem} by detailing the class of high-dimensional systems we consider and posing the associated nonlinear optimal control problem.
In Section \ref{sec:SSMLearn}, we summarize relevant results from SSM theory and outline the data-driven procedure for learning control dynamics on SSMs.
We then discuss our proposed control procedure in Section \ref{sec:MPC} and present simulation results in Section \ref{sec:results}.

\section{Problem Formulation}
\label{sec:problem}
\subsection{High-Dimensional Optimal Control Problem}
\begin{figure*}
  \includegraphics[width=\textwidth,height=5cm]{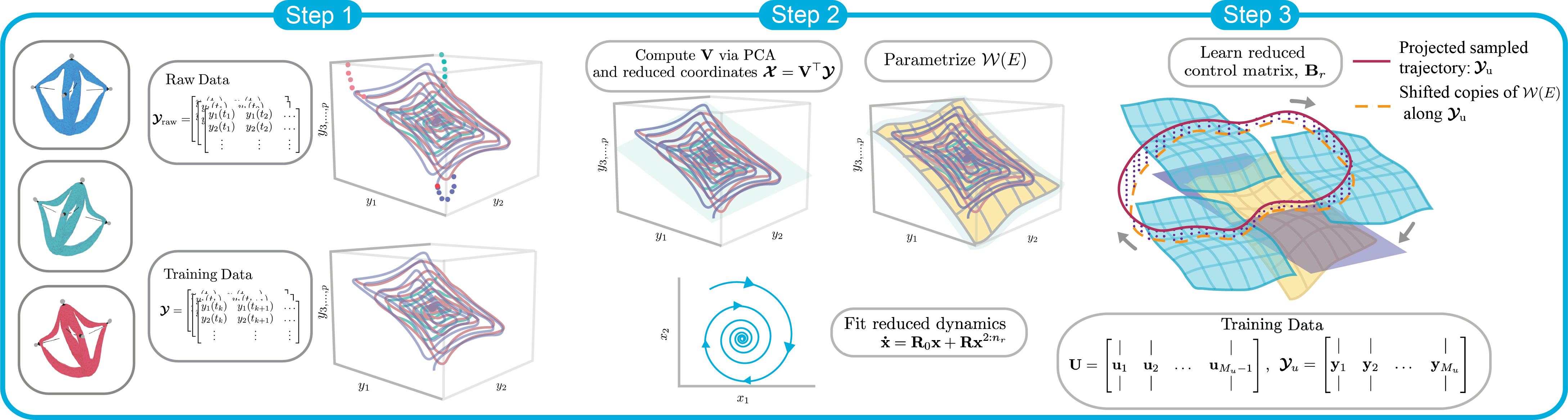}
  \caption{Three-step procedure to learn control-oriented dynamics on the SSM. Step 1 depicts the data collection procedure whereby we displace the robot across various parts of its workspace and collect decaying trajectories. We then form our training data by truncating the dataset to approximate trajectories that are on or near the manifold. Step 2 computes the SSM parameterization and autonomous dynamics while Step 3 regresses the control matrix which best explains how the autonomous SSM is translated under the influence of control. The  ``Diamond" soft robot is shown in its various displaced configurations on the far left.}
  \vspace{-0.6cm}
  \label{fig:SSMLearn}
\end{figure*}
We consider control-affine mechanical systems with $\nodes \in \N$ DOF. These systems encompass a wide range of robots such as manipulators, drones, and highly-articulated robots. In the continuum limit (\ie $\nodes \rightarrow \infty$), these systems can also converge to the exact model of control-affine soft robots \cite{della2021model}. Such systems can be written in first-order form with state vector $\fstate(t) \in \R^{\nfstate}$ (where $f$ denotes \emph{full} state, as opposed to the \emph{reduced} state $\rstate$ introduced in Section~\ref{subsec:ROMonSSM}) as
\begin{equation}
\label{eq:firstorder}
    \dot{\mathbf{x}}^f(t) = \mathbf{A} \fstate (t) + \mathbf{f}_{\text{nl}}(\fstate (t)) + \epsilon \mathbf{B} \ctrl (t),
\end{equation}
where $\nfstate = 2 \nodes$, $\mathbf{A} \in \R^{\nfstate \times \nfstate}$ is assumed to be negative-definite (\ie the origin is an asymptotically stable fixed point for $\epsilon=0$) and $\mathbf{B} \in \R^{\nfstate \times m}$ represents the linear control matrix. The nonlinear term $\mathbf{f}_{\text{nl}}: \R^\nfstate \rightarrow \R^\nfstate$ belongs to the class of analytic functions and satisfies $\mathbf{f}_{\text{nl}}(\mathbf{0}) = \mathbf{0}, \, \partial\mathbf{f}_{\text{nl}}(\mathbf{0})/\partial\fstate= \mathbf{0}$, while the parameter $0 < \epsilon \ll 1$ introduces our assumption that the magnitude of the control inputs should be moderate compared to the autonomous dynamics.
In this work, we consider control tasks near the vicinity of the robot's equilibrium point (\eg a highly-articulated manipulator arm conducting pick and place tasks in a constrained workspace). Thus, if the desired trajectories are reasonable, this assumption is typically satisfied.
Derivation of \Cref{eq:firstorder} from a second-order mechanical system can be found in Appendix~\ref{appendix:2ndorder}\iftoggle{ext}{}{ \cite{alora2022extended}}.

We now pose the problem of controlling \Cref{eq:firstorder} to follow arbitrary and dynamic trajectories in the vicinity of the origin.
Consider the following continuous-time, optimal control problem (OCP) with quadratic cost and polytopic constraints in states and control:
\begin{align}
\label{eq:OCP}
    \minimize_{\ctrl(\cdot)}~
    &\norm{\delta \perf(t_f)}^2_{\Qf} + \int_{t_0}^{t_f} \Big(\norm{\delta \perf(t)}^2_\Q + \norm{\ctrl(t)}^2_\Rcost \Big) dt, \nonumber \\
    \subjectto~& \fstate(0) = \mathbf{g}(\perf(0)), \nonumber \\
    &\dot{\mathbf{x}}^f(t) = \mathbf{A} \fstate (t) + \mathbf{f}_{\text{nl}}(\fstate (t)) + \epsilon \mathbf{B} \ctrl (t), \\
    &\obs(t) = \mathbf{h}(\fstate(t)), \quad \perf(t) = \mathbf{C}\mathbf{\obs}(t), \nonumber \\
    & \ctrl \in \Cntrl, \quad \perf \in \Observable \nonumber.
\end{align}
Here, $\delta \perf(t) = \perf(t) - \perfref(t)$ is the tracking difference between the performance variable, $\perf(t) \in \R^{\nperf}$ and the desired trajectory $\perfref(t) \in \R^\nperf$. The observed state is denoted as $\obs(t) \in \R^{\nout}$ and $[t_0, t_f]$ represents the time horizon. 
$\Q, \Qf \in \R^{\nperf \times \nperf}$ are positive semi-definite matrices which represent the stage and terminal costs, respectively, over the performance variables, while $\Rcost$ is a positive-definite matrix representing the cost on controls. 
The functions $\mathbf{g}: \R^\nperf \rightarrow \R^\nfstate$ and $\mathbf{h}: \R^\nfstate \rightarrow \R^\nout$ map the performance variable to the full state and the full state to the observed state, respectively, while $\mathbf{C} \in \R^{\nperf \times \nout}$ is a selection matrix of states that we observe. 
Lastly, the constraint sets are defined as $\Cntrl := \{ \ctrl(t) \in \R^\nctrl : \mathbf{M}_{\mathrm{u}} \ctrl(t) \leq \mathbf{b}_{\mathrm{u}} \}$ and $\Observable := \{ \perf \in \R^\nperf : \mathbf{M}_{\mathrm{z}} \perf(t) \leq \mathbf{b}_{\mathrm{z}} \}$  with $\mathbf{M}_{\mathrm{u}} \in \R^{n_u \times m}$ and $\mathbf{M}_{\mathrm{z}} \in \R^{n_z \times o}$, where $n_u$ and $n_z$ represent the number of constraints in the inputs and the observed states, respectively. 

For high-dimensional dynamical systems, \ie $n_f \gg 1$, dimensionality of \Cref{eq:firstorder} becomes a bottleneck and it is intractable to solve  the OCP~(\ref{eq:OCP}) in an online fashion.
Thus, we seek a low-dimensional approximation of \Cref{eq:firstorder} that enables online control and allows us to approximate a solution to the OCP.

\section{Data-Driven Modeling of \\ Low-dimensional Dynamics}
\label{sec:SSMLearn}
In this section, we describe our data-driven SSMR procedure to construct controlled, predictive models of soft robots from data. 
Our approach entails learning low-dimensional models directly as the reduced dynamics on attracting, low-dimensional invariant manifolds that generically exist in dissipative physical systems.

\subsection{SSMs in a Nutshell}
We define an $\nrstate$-dimensional spectral subspace $\spectralsubspace$ as the direct sum of an arbitrary collection of $\nrstate$ eigenspaces of $\mathbf{A}$,
\begin{align*}
    E \defeq  E_{j_1} \oplus E_{j_2} \oplus ... \oplus E_{j_\nrstate},
\end{align*}
where $\spectralsubspace_{j_k}$ denotes the real eigenspace corresponding to an eigenvalue $\lambda_{j_k}$ of $\mathbf{A}$. Let $\Lambda_{\spectralsubspace}$ be the set of eigenvalues related to $\spectralsubspace$ and $\Lambda_{\text{out}}$ be that of eigenvalues not related to $\spectralsubspace$.
If $\min_{\lambda\in\Lambda_{\spectralsubspace}} \text{Re}(\lambda)>\max_{\lambda\in\Lambda_{\text{out}}} \text{Re}(\lambda)$, then $\spectralsubspace$ represents the slowest spectral subspace of order $\nrstate$.
Intuitively, the slowest spectral subspace corresponds to the dominant modes representing the persisting dynamics of the robot and can be extracted via modal analysis or principal component analysis (PCA).

Let us first assume that $\epsilon = 0$. For purely linear systems without external forcing (\ie the linearization of \Cref{eq:firstorder}), any trajectories that start in $\spectralsubspace$ will remain in $\spectralsubspace$, by the Spectral Mapping Theorem.
When nonlinearities are introduced, superposition is lost and the autonomous part of \Cref{eq:firstorder} is no longer invariant on $\spectralsubspace$.
The autonomous SSM corresponding to $\spectralsubspace$, $\autSSM$, is the smoothest $n$-dimensional manifold in the robot's phase space which nonlinearly extends the invariance of $\spectralsubspace$, \ie for the autonomous part of \Cref{eq:firstorder}
\begin{align}
    \fstate_{\mathrm{aut}} (0) \in \autSSM \implies \fstate_{\mathrm{aut}}(t) \in \autSSM, \quad \forall t \in \R,
\end{align}
where $\dot{\mathbf{x}}^f_{\mathrm{aut}}(t) = \mathbf{A} \fstate_{\mathrm{aut}} (t) + \mathbf{f}_{\text{nl}}(\fstate_{\mathrm{aut}} (t))$ and $\fstate_{\mathrm{aut}} \in \R^{\nfstate}$.
Given the smoothness of the SSM, the parameterization of $\autSSM$ and the corresponding reduced dynamics can be represented by polynomial maps \cite{jain2021compute}, as detailed in Section~\ref{subsec:ROMonSSM}. 

Low-dimensional slow SSMs corresponding to slow spectral subspaces are ideal candidates for model reduction as nearby full system trajectories become exponentially attracted towards these manifolds and synchronize with the slow dynamics. 
Figure~\ref{fig:SSM} gives a visual depiction depiction of this property as well as the relationship between $\spectralsubspace$ and $\autSSM$.
For a detailed definition of SSMs, see Appendix~\ref{appendix:SSM}\iftoggle{ext}{}{ \cite{alora2022extended}}.

For small $\epsilon$, the SSM $\autSSM$ is still relevant for control. From a theoretical point of view, results on the existence of non-autonomous SSMs subject to quasi-periodic forcing were established in \cite{haller2016nonlinear}. Since quasi-periodic signals over a finite time interval are dense in the space of continuous signals over the same interval, we can interpret the trajectory of the system under control input as lying approximately on a time-varying, invariant manifold that is $\epsilon$-perturbed from $\autSSM$.

\subsection{Reduced-order Models on SSMs}\label{subsec:ROMonSSM}
In general, since we seldom have access to the full state $\fstate$, we must construct the SSM and the reduced dynamics of our system in the space of observed states such that $p \geq 2n + 1$ either by Whitney or Takens embedding theorems \cite{cenedese2022ssmlearn}.
In case $\obs$ does not satisfy this condition, we use time-delay embeddings of $\obs$, whereby our new observed measurements include current and past measurements of $\obs$, in order to embed $\autSSM$ in a space with sufficient dimension.

To describe the geometry of $\autSSM$, we seek a pair $\mathbf{w}(\rstate)$, $\mathbf{v}(\obs)$ of smooth, invertible functions where $\obs = \xtoy(\rstate)$ uniquely maps the reduced state on the SSM to the observed state and $\rstate = \ytox(\obs)$ maps the observed state to the reduced coordinates, where $\rstate \in \R^{\nrstate}$ is the reduced state. 
By definition of the invariance and tangency properties of the SSM, the two maps that parameterize $\autSSM$ must satisfy the invertibility relations \cite{haller2016nonlinear}, $\obs = (\xtoy \circ \ytox)(\obs)$ and $\rstate = (\ytox \circ \xtoy)(\rstate)$ such that
\begin{align}
\label{eq:parameterization}
    & \rstate = \ytox(\obs) \defeq \mathbf{V}^\top \obs, \nonumber \\
    & \obs = \xtoy(\rstate) \defeq \mathbf{W}_0 \rstate + \mathbf{W} \rstate^{2:n_w},
\end{align}
where $\rstate^{2:n_w}$ is the family of all monomials from order $2$ to $n_w$, and $n_w$ is the desired order of the Taylor series expansion for approximating the SSM.
Also, the columns of $\mathbf{V} \in \R^{\nfstate \times \nrstate}$ span the spectral subspace of $\spectralsubspace$ and $\mathbf{W}_0, \mathbf{W}$ represent coefficient matrices of the SSM parameterization. 
In addition, the reduced dynamics on $\autSSM$ are represented by
\begin{align}
\label{eq:ROMaut}
    \dot{\rstate}_{\text{aut}} = \mathbf{r}_{\text{aut}}(\rstate) \defeq \mathbf{R}_0 \rstate + \mathbf{R} \rstate^{2:n_r},
\end{align}
where $\mathbf{r}_{\text{aut}}: \R^\nrstate \rightarrow \R^\nrstate$ is the autonomous reduced dynamics on the SSM; $\mathbf{R}_0$ and $\mathbf{R}$ represent the corresponding coefficient matrices. 
Since $\autSSM$ is locally a graph over the spectral subspace $E$ \cite{cenedese2022ssmlearn}, we can identify the full state trajectory of System~(\ref{eq:firstorder}) on $\autSSM$ described by \Cref{eq:ROMaut}. Figure~\ref{fig:SSM} gives an intuitive depiction of this concept.

We seek to learn the SSM-reduced dynamics for control and construct mappings that describe the trajectory of our observed states on SSMs.
Our three-step SSMR procedure involves: (1) collecting trajectories at or near the SSM. (2) learning the SSM geometry and the reduced dynamics in \Cref{eq:ROMaut}, followed by (3) learning a linear control matrix that describes the effect of the controls in the reduced coordinates. Figure~\ref{fig:SSMLearn} summarizes the complete data-driven SSMR approach.

\subsection{Learning Autonomous Dynamics on SSMs}

To learn the geometry and reduced dynamics on the SSM, the training data should involve only trajectories that are near the SSM.
Thus, we obtain training data snapshots by displacing the robot along various directions in its workspace, then collect the observed state trajectory as it decays to its equilibrium position. In other words, we form an augmented matrix of $M_y$ (possibly time-delayed) decay datasets $\Ydata_{\mathrm{raw}} = \Big[ \Ydata_1, \dots, \Ydata_{M_y} \Big]$, as shown in Figure~\ref{fig:SSMLearn} (left). We remove initial transients converging to the SSM in our datasets $\Ydata_{\mathrm{raw}}$ by truncating the first few states in the decay trajectories \cite{cenedese2022data} and forming the dataset, $\Ydata$.

To start, we first compute $\mathbf{V}$ by finding the $n$ dominant modes of \Cref{eq:firstorder}. 
To do this we carry out principle component analysis on the trajectory dataset $\Ydata$ and pick the $n$ leading directions that capture a majority of the variance in the data.
Indeed, for systems that do not feature strong nonlinearities, PCA is able to obtain a close estimate for the spectral subspace $\spectralsubspace$ to which the SSM is tangent \cite{axaas2022fast}.

Once we project $\Ydata$ onto the reduced coordinate such that $\X = \mathbf{V}^\top \Ydata$, we can then learn the parameterization of $\autSSM$ (\ie learn the map $\mathbf{w}$) by finding $\mathbf{W}$ and $\mathbf{W}_0$ via polynomial regression
\begin{align}
\label{eq:maps}
    (\mathbf{W}_0^*, \mathbf{W}^*) = & \argmin_{\mathbf{W}_0, \mathbf{W}} \norm{\Ydata - \mathbf{W}_0 \X - \mathbf{W}\X^{2:n_w}}^2_F.
\end{align}
In a similar fashion, we can compute the polynomial form of the reduced dynamics in \Cref{eq:ROM} by finding the coefficients $\mathbf{R}_0$ and $\mathbf{R}$ via the regression
\begin{align}
\label{eq:learnROM}
    (\mathbf{R}_0^*, \mathbf{R}^*) = \argmin_{\mathbf{R}_0, \mathbf{R}} \norm{\dot{\X} - \mathbf{R}_0 \X - \mathbf{R} \X^{2:n_r}}^2_F,
\end{align}
The time derivative in \Cref{eq:learnROM} can be computed using standard finite difference schemes if the sampling time of $\X$ is much smaller than the Nyquist sampling time of the fastest mode in the SSM dynamics.
Otherwise, we can also compute a discrete-time alternative to \Cref{eq:ROM} using a similar procedure through simple shifting operations on the dataset as in \cite{proctor2016dynamic}.

The procedure outlined above is suitable for both numerical and experimental data. For the former, we can choose $\obs (t) = \fstate (t)$ and let the dataset $\Ydata$ consist of full state information during decay. We implement this procedure using a modified version of \textit{SSMLearn}~\footnote[1]{https://github.com/StanfordASL/SSMR-for-control} \cite{cenedese2022ssmlearn}.

\subsection{Learning the Control Matrix}
Keeping in mind our moderate control assumption, we assume $\epsilon = 1$, without loss of generality, for the rest of the exposition. Once the reduced autonomous dynamics on $\autSSM$ is known, we seek to learn the contribution of control in the reduced coordinates.
Our goal is to find the best linear control matrix $\Br \in \R^{\nrstate \times \nctrl}$ which best explains the difference between the controlled dynamics and our model of the autonomous dynamics. 
We explore the actuation space of the robot by randomly sampling a sequence of inputs, $\U$, and recording the corresponding (possibly time-delayed) observed state trajectory $\Ydata_{u}$, as depicted in Figure~\ref{fig:SSMLearn} (right).
\begin{figure*}
\includegraphics[width=\textwidth]{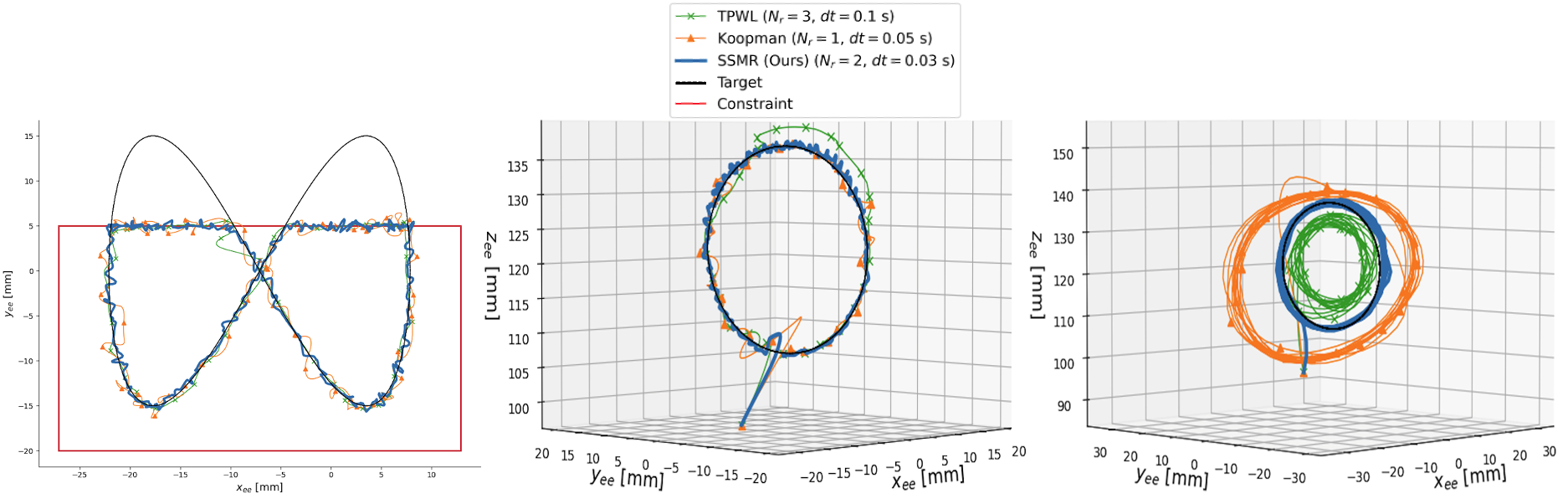}
  \caption{Simulation results of tracking performance for tasks (1), (2), and (3) from left to right, respectively, with horizon length of $N = 3$. The parameters used were tuned for each method to yield the best, real-time performance across all tasks. The TPWL trajectory is shown in green, the Koopman trajectory in orange, and the SSMR in blue. The dotted black line represents the reference trajectory while the red line represent constraints. The quasi-static circle (task 2) MSEs (in mm$^2$) are $e_{\text{TPWL}} = 3.35$, $e_{\text{koop}} = 0.91$, and $e_{\mathrm{SSM}} = 0.53$. The near-resonance circle (task 3) MSEs are $e_{\mathrm{TPWL}} = 21.75$, $e_{\text{koop}} = 133.6$, and $e_{\mathrm{SSM}} = 1.87$.}
  \label{fig:figure8}
  \vspace{-0.5cm}
\end{figure*}

We then project the observed states down to the reduced coordinates and form the reduced state matrix $\X_{\mathrm{u}} = \mathbf{V}^\top \Ydata_{\text{u}}$. 
Additionally, we evaluate our model of the autonomous dynamics and form the matrix $\dot{\X}_{\text{aut}}$ $= \mathbf{r}_{\mathrm{aut}} (\X_{\text{u}})$. 
Learning the (continuous-time) control matrix from data amounts to solving the minimization problem
\begin{align}
\label{eq:learnB}
    \Br^* = \argmin_{\Br} \norm{\dot{\X}_{\mathrm{u}} - \dot{\X}_{\mathrm{aut}} - \Br \U}^2_F,
\end{align}
where $\dot{\X}_\text{u}$ is computed by finite differencing $\X_\text{u}$. Our learned, low-dimensional control dynamics is thus,
\begin{align}
\label{eq:ROM}
    \dot{\rstate} = \mathbf{r}(\rstate, \ctrl) \defeq \mathbf{R}_0 \rstate + \mathbf{R} \rstate^{2:n_r} + \Br \ctrl.
\end{align}

In general, the introduction of control causes $\autSSM$ to lose its invariance.
Intuitively, though, we expect that the trajectories will remain within a small neighborhood of $\autSSM$ since the effect of our control input is moderate compared to the system dynamics. 
Thus, we interpret this step as regressing a linear matrix that optimally translates the autonomous SSM under control inputs to be as close as possible to off-SSM trajectories.
\section{SSM-based Nonlinear MPC}
\label{sec:MPC}
\subsection{Reduced Order Optimal Control Problem}
Learning the parameterization of $\autSSM$ enables us to learn the intrinsic physics of our system, leading to low-dimensional and accurate reduced models with $\nrstate \ll \nfstate$.
This allows us to approximate the OCP in \eqref{eq:OCP} by posing an optimization problem with respect to the dynamics on the SSM as follows 
\begin{align}
\label{eq:ROMOCP}
        \minimize_{\mathbf{u(\cdot)}}~
        & \norm{\delta \perf(t_f)}^2_{\mathbf{Q}_f} + \sum_{k=1}^{N-1} \Big( \norm{\delta \perf(t)}^2_{\mathbf{Q}} + \norm{\mathbf{u}(t)}^2_{\mathbf{R}}\Big)
 \nonumber \\
    \mathrm{subject~to}~
        &\mathbf{\rstate}(0) = \mathbf{V}^\top(\mathbf{y}(0) - \mathbf{y}_{\text{eq}}), \nonumber \\
        &\mathbf{\dot{\rstate}}(t) = \mathbf{r}(\mathbf{\rstate}(t)) +  \mathbf{B}_r \ctrl(t), \nonumber\\
        &\perf(t) = \mathbf{C} \mathbf{w}(\mathbf{\rstate}(t)) + \mathbf{z}_{\text{eq}}, \\
        & \mathbf{z}(t) \in \Observable, \quad \ctrl(t) \in \Cntrl, \nonumber
\end{align}
where $\perf_{\text{eq}} \in \R^\nperf$ and $\obs_{\text{eq}} \in \R^\nout$ are the performance and observed states at equilibrium.
To solve the approximate OCP \eqref{eq:ROMOCP} numerically, we discretize the continuous-time system
and use Sequential Convex Programming (SCP) to transform \eqref{eq:ROMOCP} into a sequence of quadratic programs. 
If $\nrstate$ is small enough, we can compute the solution to the resulting approximate OCP in real-time. We implement our SSMR-based controller on top of the open-source soft robot control library\footnote[2]{https://github.com/StanfordASL/soft-robot-control} presented in ~\cite{tonkens2021soft}. See Appendix~\ref{appendix:SCP}\iftoggle{ext}{}{ \cite{alora2022extended}} for more details on the SCP setup.

\section{Simulation Results}
\label{sec:results}
\begin{figure}
    \centering
    \includegraphics[width=\linewidth]{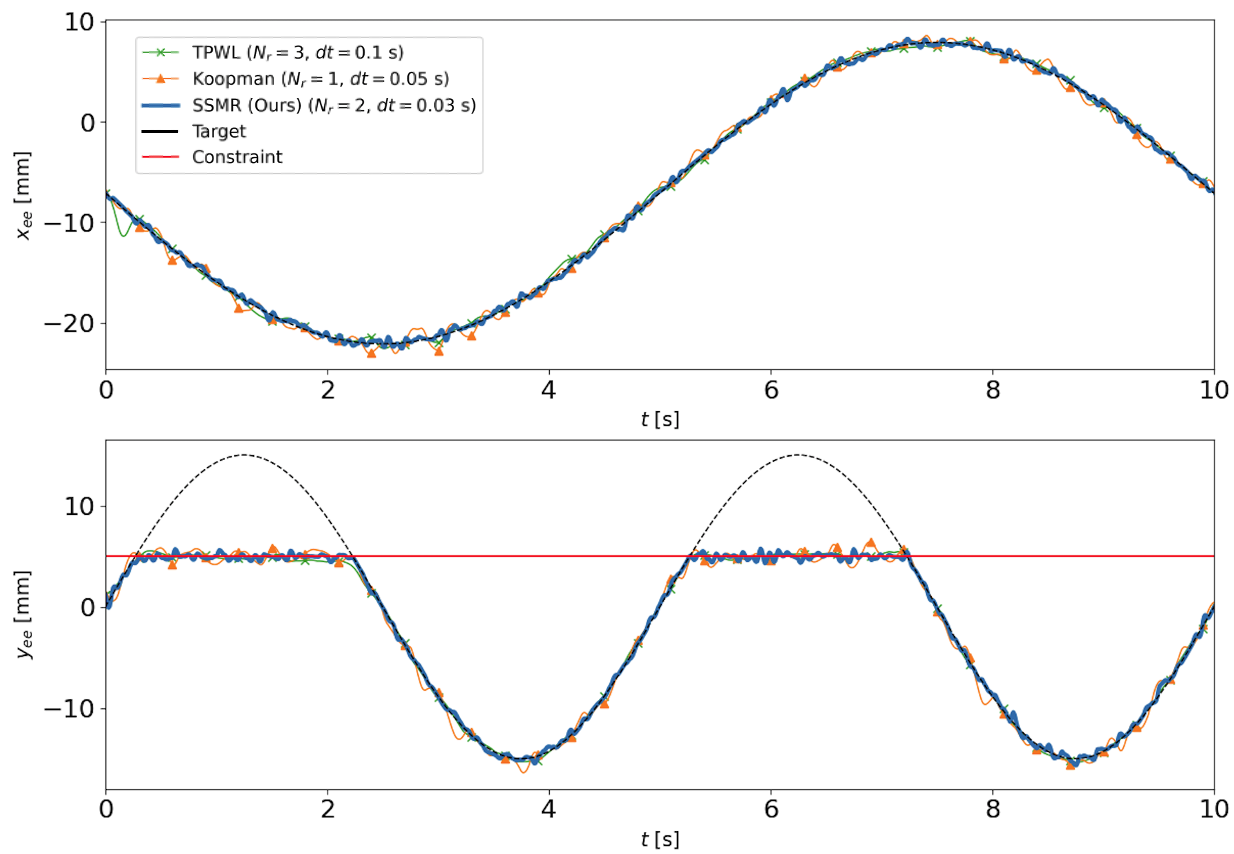}
    \caption{Time-series simulation results of tracking performance for the quasi-static Figure Eight (task 1). The controller parameters for each approach are set similarly to those reported in Figure~\ref{fig:figure8}. The MSE (in mm$^2$) for the TPWL, Koopman, and SSMR approaches are $e_{\mathrm{TPWL}} = 0.22$, $e_{\mathrm{koop}} = 0.38$, and $e_{\mathrm{SSM}} = 0.13$.}
    \label{fig:circle}
    \vspace{-0.5cm}
\end{figure}
\subsection{Simulation}
We now compare our proposed SSMR method against the Trajectory Piecewise-Linear (TPWL) approach~\cite{tonkens2021soft} and Koopman operator-based control approach~\cite{bruder2019modeling} in simulations of the elastomer ``Diamond" robot (shown in Figure~\ref{fig:SSMLearn}), as detailed in \cite{tonkens2021soft} and \cite{Lorenzetti2021}. 
We show that our approach outperforms these baselines while mitigating their drawbacks, namely lack of generalization, lack of robustness to noise, and computational intractability.
\begin{table*}
\caption{The table on the left shows mean squared error ($mm^2$) for all considered trajectories while the right shows average cumulative QP solve times (in milliseconds) for the SCP algorithm. The Koopman model consists of polynomials up to order 2 over $\perf = [x_{\text{ee}}, y_{\text{ee}}, z_{\text{ee}}]^\top$ with a single time delay ($d_{\text{koop}} = 66$). We train separate, discrete-time Koopman models corresponding to the various time-discretizations, $dt$. The TPWL model parameters are set similarly to the ones reported in \cite{Lorenzetti2021} ($d_{\text{TPWL}} = 42$). We learn a single, continuous-time SSM model of cubic order and although it is low-dimensional ($d_{\text{SSM}} = 6$), it outperforms the other approaches in both tracking performance (at low enough time-discretization) and solve time. The QP is solved using Gurobi~\cite{gurobi} on a 1.6 GHz Intel Core i5 processor with 8 GB of RAM.}
\begin{adjustbox}{width=\textwidth}
\begin{tabular}{c | c}\toprule
\begin{tabular}{c|c|c}
\multicolumn{1}{c}{\parbox[t]{2in}{\hspace*{1.2in}{\textbf{Figure Eight}}}} & \multicolumn{1}{c}{\textbf{Circle}} & \multicolumn{1}{c}{\textbf{Near-Resonance Circle}} \\
\begin{tabular}{p{0.0001cm} c} \midrule
&
\begin{tabular}{@{}p{2.7cm} p{0.6cm}p{0.6cm}p{0.7cm}p{0.8cm}@{}}
\raggedleft $dt$ (ms) & \raggedleft $\mathit{10}$ & \raggedleft $\mathit{20}$ & \raggedleft $\mathit{50}$ &\raggedleft $\mathit{100}$
\end{tabular} \\
\midrule
\rotatebox{90}{\parbox[][0.1cm][t]{0.2in}{\hspace*{-.12in}{$\mathbf{N = 3}$}\hspace*{\fill}}} &
\begin{tabular}{|@{}p{2.7cm} p{0.7cm}p{0.7cm}p{0.7cm}p{0.7cm}@{}}
\raggedleft SSMR (Ours) & \raggedleft \textbf{0.131} & \raggedleft \textbf{0.123} & \raggedleft 0.227 & 0.718 \\

\raggedleft TPWL & 0.166 & 0.149 & \textbf{0.164} & \textbf{0.191}  \\

\raggedleft Koopman EDMD & 1.053 & 1.230 & 0.500 & 0.725
\end{tabular} \\
\midrule
\rotatebox{90}{\parbox[][0.1cm][t]{0.2in}{\hspace*{-.12in}{$\mathbf{N = 5}$}\hspace*{\fill}}} &
\begin{tabular}{|@{}p{2.7cm} p{0.7cm}p{0.7cm}p{0.7cm}p{0.7cm}@{}}
\raggedleft SSMR (Ours) & \raggedleft \textbf{0.153} & \raggedleft \textbf{0.136} & \raggedleft 0.205 & 0.757 \\

\raggedleft TPWL & 0.160 & 0.155 & \textbf{0.160} & \textbf{0.196}  \\

\raggedleft Koopman EDMD & 1.540 & 1.286 & 0.515 & 0.679
\end{tabular} \\

\end{tabular} & 
\begin{tabular}{c} \midrule
\begin{tabular}{p{0.6cm}p{0.6cm}p{0.7cm}p{0.8cm}}
\raggedleft $\mathit{10}$ & \raggedleft $\mathit{20}$ & \raggedleft $\mathit{50}$ &\raggedleft $\mathit{100}$
\end{tabular} \\
\midrule
\begin{tabular}{p{0.7cm}p{0.7cm}p{0.7cm}p{0.7cm}}
\raggedleft \textbf{0.481} & \raggedleft \textbf{0.342} & \raggedleft 1.353 & 4.480 \\

3.996 & 3.033 & 3.197 & 3.216  \\

2.985 & 1.806 & \textbf{0.914} & \textbf{2.060}
\end{tabular} \\
\midrule
\begin{tabular}{p{0.7cm}p{0.7cm}p{0.7cm}p{0.7cm}}
\raggedleft \textbf{0.466} & \raggedleft \textbf{0.348} & \raggedleft 1.287 & 4.561 \\

3.350 & 3.278 & 3.265 & 3.254  \\

2.919 & 1.632 & \textbf{0.895} & \textbf{2.119}
\end{tabular} \\

\end{tabular} &
\begin{tabular}{c} \midrule
\begin{tabular}{p{0.7cm}p{0.5cm}p{0.7cm}}
$\mathit{10}$ & $\mathit{20}$ & \raggedleft $\mathit{50}$ 
\end{tabular} \\
\midrule
\begin{tabular}{ccc}
\textbf{0.893} & \textbf{1.861} & 32.87 \\

4.077 & 3.789 & \textbf{4.472} \\

1.864 & 6.707 & 135.3
\end{tabular} \\
\midrule
\begin{tabular}{ccc}
\textbf{0.810} & \textbf{1.816} & 33.43 \\

3.698 & 3.900 & \textbf{4.585}  \\

1.818 & 7.149 & 133.6
\end{tabular} \\
\end{tabular} \\
\end{tabular} & 
\begin{tabular}{c}
\textbf{Average Solve Times (ms)} \\

\begin{tabular}{c} \midrule
\begin{tabular}{p{0.6cm}p{0.4cm}p{0.7cm}p{0.7cm}}
$\mathit{10}$ & $\mathit{20}$ & \raggedleft $\mathit{50}$ &\raggedleft $\mathit{100}$
\end{tabular} \\
\midrule
\begin{tabular}{p{0.7cm}p{0.7cm}p{0.7cm}p{0.7cm}}
\raggedleft \textbf{0.85} & \raggedleft \textbf{0.97} & \raggedleft \textbf{0.97} & \textbf{0.92} \\

25.31 & 26.19 & 27.75 & 31.32  \\

6.08 & 6.10 & 5.95 & 5.92
\end{tabular} \\
\midrule
\begin{tabular}{p{0.7cm}p{0.7cm}p{0.7cm}p{0.7cm}}
\raggedleft \textbf{1.55} & \raggedleft \textbf{1.49} & \raggedleft \textbf{1.62} & \textbf{1.26} \\

52.20 & 51.23 & 55.51 & 58.52  \\

15.81 & 16.18 & 18.01 & 19.65
\end{tabular} \\
\end{tabular} \\
\end{tabular} \\
\bottomrule
\end{tabular}
\end{adjustbox}
\label{table:mseandtimes}
\vspace{-0.5cm}
\end{table*}

We carry out simulations using the finite-element based SOFA framework \cite{allard2007sofa}; the Diamond robot mesh we used for simulation can be found in the \textit{SoftRobots} plugin \cite{coevoet2017software}. 
The parameters of our Diamond robot mirror those reported for a hardware replica in \cite{Lorenzetti2021}, where $E = 175$ MPa is the Young's modulus, $\nu = 0.45$ is the Poisson ratio, and $\alpha = 2.5, \beta = 0.01$ represent the usual parameters for Rayleigh damping \ie $C = \alpha \mathbf{M} + \beta \mathbf{K}$. 
Additionally, $N_r \leq N$ represents the rollout horizon of the optimal solution, $\ctrl^*$, to OCP~\eqref{eq:ROMOCP} while the controller sampling time is $T_c = N_r dt$, where $dt$ is the time-discretization of the dynamics.


In this work, we consider control tasks in which the end effector of the robot is made to follow various trajectories.
Thus, the performance variable $\perf = [x_{\text{ee}}, y_{\text{ee}}, z_{\text{ee}}]^\top$ denotes the position of the top of the robot in its workspace.
We also introduce additive Gaussian measurement noise to simulate real-world conditions.
We consider three control tasks which include following (1) a figure eight in the x--y plane subject to constraints, (2) a circle in the y--z plane, and (3) the same circle but near resonance with the dominant mode of the system. 

Since we are in a simulation environment, we collect full state information as training data \ie the $i$-th dataset is $\Ydata_i = [\fstate_1, \fstate_2, \dots \fstate_M]^\top$. 
We obtain this data by displacing the robot along $44$ different points in its workspace and observe the decaying trajectory state transitions sampled at $T_s = 1$ ms. This is consistent with the highest frequency mode in the SSM which has a period of roughly 330 ms.
After conducting PCA on our training data, we found that the 3 leading configuration modes (6 modes in phase space) captured more than $95$\% of the variance in our dataset.
Hence, we learn a cubic order, 6-D autonomous SSM parametrization described in \eqref{eq:maps} and its continuous-time, reduced dynamics \eqref{eq:ROM} using the procedure in Section~\ref{sec:SSMLearn}.
Lastly, we learn the control matrix by randomly sampling controls and then collect the resulting state transitions sampled at $10$ ms.

Table~\ref{table:mseandtimes} reports the mean-squared error tracking performance and average \textit{cumulative} time to solve the QP for all trajectories at various controller parameters and time discretization of \Cref{eq:ROM}.
To enable real-time control, we seek control parameters such that the controller sampling time is at least an order of magnitude less than the solve time.
Figures~\ref{fig:figure8} and \ref{fig:circle} depict simulation results for trajectories (1), (2), and (3) for controller parameters chosen to maximize performance while enabling real-time control.

These results show that our SSMR-based MPC scheme outperforms the TPWL and the Koopman approach in tracking performance across all trajectories considered, for small enough time discretization.
Thus, our approach exhibits superior generalizability to control tasks as shown in the above figures and tables.
Due to the low dimensionality of our learned model, we can solve the SCP iterations quickly and the computational burden grows modestly as the MPC horizon increases. 
As shown in Table~\ref{table:mseandtimes}, the solve times for our approach are magnitudes lower than for the TPWL and Koopman-based methods, giving us more freedom to choose the controller parameters to enable real-time control.
Observed deterioration of performance as $dt$ increases in \Cref{table:mseandtimes} is likely due to numerical errors introduced by coarser time-discretization of the dynamics since we learn a continuous-time model of \Cref{eq:ROM}.

\subsection{Discussion}
Additionally, the SSMR approach offers several practical advantages over the alternatives.
First, our SSM-based model exhibits good closed-loop performance at longer horizons and does not suffer from numerical conditioning issues that plague the Koopman approach.
We found that at horizons $N \geq 10$, the Koopman QPs were no longer solvable, which is likely due to ill-condition of the Koopman matrices.
It is well-known that approximation of the Koopman operator is numerically challenging when many observables are considered \cite{dahdah2022system}. Since we explicitly reason about the dynamics of the system in the learning process, we find that SSMR yields radically low-dimensional and thus, numerically well-behaved, models.

Second, our approach involves only two parameters: the order approximation and dimension of the manifold.
Of these two, the dimension of the SSM is a property of the system dynamics, which can be inferred via a frequency analysis of the available data. The polynomial order of the SSM approximation controls the accuracy and the trade-off between generalization and overfitting.
The size of the Koopman model grows rapidly with the number of observed states while the dimension of the projection basis for TPWL needs to be fairly large for acceptable closed-loop performance.
In contrast, since off-manifold dynamics are sufficiently approximated by those on the SSM for closed-loop control, we can learn models of minimal size and tune the SSM order iteratively to increase model fidelity, as needed.
This has considerable practical advantage over learning-based approaches where it is well-known that closed-loop performance is highly sensitive to choice of dictionary features, size, and regularization.


\section{Conclusion and Future Work}
In this work, we proposed a new data-driven approach for constructing control-oriented, reduced models of soft robots on spectral submanifolds. 
Using our approach, we can construct faithful, predictive, low-dimensional models which can be effectively used for real-time optimal control.
We demonstrated that our SSM-based MPC scheme outperforms the state of the art significantly in both tracking error and computation time. 
The success of \textit{SSMLearn} \cite{cenedese2022data} in the experimental domain hints at the prospects of our SSMR approach for application to real world robots.
Bolstered by promising results in a high-fidelity, finite-element simulation environment, we plan to validate the data-driven SSMR approach on our hardware platform detailed in \cite{Lorenzetti2021}.

While these results are promising, there are many open problems. For example, although the setting we consider involves tasks around an equilibrium point (\eg manipulation tasks in a constrained workspace), many high-dimensional systems are not fixed to a point and can freely navigate their environment. Extending our SSMR framework to handle these settings would generalize our approach to a broader class of systems.
Also, since most robotic systems have configuration-dependent actuation constraints, we plan to extend our approach to learning dynamics with state-affine control.
Lastly, we plan to estimate errors arising from SSM approximation \textit{a-priori} and derive error bounds for constraint-tightening control schemes in an MPC framework.

\addtolength{\textheight}{0cm}   

\vspace{0.5em}\noindent\emph{Acknowledgements.}
The authors thank Elisabeth Alora and Matteo Zallio for generating the instructive figures, Florian Mahlknecht for thoughtful discussions and help with initial implementations, and Spencer M.\ Richards for his careful review of the manuscript.

\printbibliography

\iftoggle{ext}{
\appendix
\subsection{High-Dimensional Mechanical System}
\label{appendix:2ndorder}
We consider robots modeled as mechanical systems with second-order form
\begin{equation}
\label{eq:2ndorder}
    \mass \ddot{\q}(t) + \damp \dot{\q}(t) + \stiff \q(t) + \mathbf{F}_{\text{int}}(\q(t), \dot{\q}(t)) = \mathbf{H} \ctrl(t),
\end{equation}
where $\q(t) \in \R^{\nodes}$ is the vector of generalized coordinates, $\mass \in \R^{\nodes \times \nodes}$ is the mass matrix, $\mathbf{C} \in \R^{\nodes \times \nodes}$ represents the damping matrix, $\stiff \in \R^{\nodes \times \nodes}$ is the stiffness matrix, and $\mathbf{F}_{\text{int}}(\q, \dot{\q}) \in \R^{\nodes}$ represents the internal nonlinear forces. Also $\ctrl(t) \in \R^{m}$ is the vector of inputs and $\mathbf{H} \in \R^{\nodes \times m}$ represents the linear mapping of actuation forces from their point of application to the configuration space.

\Cref{eq:2ndorder} can be written in the first-order form of \Cref{eq:firstorder} with the state vector $\fstate(t) = [\q(t), \dot{\q}(t)]^\top \in \R^{\nfstate}$ such that
\begin{align}
    \fstate = \begin{bmatrix}\mathbf{q} \\ \dot{\mathbf{q}} \end{bmatrix}, \quad
    \mathbf{A} = \begin{bmatrix}0 & \mathbf{I} \\
                    -\mathbf{M}^{-1} \mathbf{K} & -\mathbf{M}^{-1} \mathbf{C}\end{bmatrix}, \nonumber \\
    \mathbf{f}_{\mathrm{nl}}(\fstate) = \begin{bmatrix} 0 \\ -\mathbf{M}^{-1} \mathbf{F}_{\mathrm{int}}(\fstate) \end{bmatrix}, \quad \mathbf{B} = \begin{bmatrix} 0 \\ \mathbf{V}^\top \mathbf{M}^{-1} \mathbf{H} \end{bmatrix}.
\end{align}

\subsection{Spectral Submanifolds}
\label{appendix:SSM}
Recent results in nonlinear dynamics establish the existence of unique, smoothest invariant structures in the phase space of \Cref{eq:firstorder} \cite{haller2016nonlinear}. 
SSMs are nonlinear continuations of the spectral subspaces of the linearization of \Cref{eq:firstorder}. 
The SSM corresponding to $\spectralsubspace$ in the autonomous part of \Cref{eq:firstorder} is defined as follows.
\defn{An autonomous SSM $\autSSM$, corresponding to a spectral subspace $E$ of the operator $\mathbf{A}$ is an invariant manifold of the autonomous part of the nonlinear system~\eqref{eq:firstorder} such that}
\begin{enumerate}
    \item $\autSSM$ is tangent to $E$ at the origin and has the same dimension as $E$,
    \item $\autSSM$ is strictly smoother than any other invariant manifold satisfying condition 1 above. 
\end{enumerate}
A slow SSM is associated with a non-resonant spectral subspace containing the slowest decaying eigenvectors of the linearized system. SSMs as described in Definition 1 turn out to exist as long as the spectrum of $\mathbf{A}\vert_E$, $\Lambda_{\spectralsubspace}$, has no low-order resonance relationship with any eigenvalue in the outer spectrum $\Lambda_{\text{out}}$ (see \cite{haller2016nonlinear, cenedese2022ssmlearn} for details).

\subsection{Convex Formulation for Real-Time Control}
\label{appendix:SCP}
Convex optimization allows us to leverage fast, iterative algorithms with polynomial-time complexity \cite{MalyutaEtAl2022} to efficiently and reliably approximate a solution to \eqref{eq:ROMOCP}.
We use sequential convex programming (SCP) to transform the nonlinear equality constraints into a sequence of linear equality constraints.
The key idea is then to iteratively re-linearize the dynamics around a nominal trajectory and solve a convex approximation near this trajectory until convergence to a local optimum of the continuous-time OCP~ \eqref{eq:ROMOCP} is observed.

To be precise, suppose we have some nominal trajectory of states and controls $(\rstate^j, \ctrl^j) = (\{ \rstate^j_k\}_{k=1}^N, \{ \ctrl^j_k\}_{k=1}^{N-1})$ at the $j$-th iteration about which we linearize the nonlinear constraints. 
We can then define the resulting linearized OCP, \textbf{(LOCP)}$_{j+1}$, as follows
\begin{align}
        \minimize_{\mathbf{u_{1:N-1}}}~
        & \norm{\delta \perf_N}^2_{\mathbf{Q}_f} + \sum_{k=1}^{N-1} \Big( \norm{\delta \perf_k}^2_{\mathbf{Q}} + \norm{\mathbf{u}_k}^2_{\mathbf{R}}\Big) \nonumber \\
    \mathrm{subject~to}~
        & \rstate_1 = \mathbf{V}^\top(\obs_1 - \obs_{\text{eq}}) \nonumber \\
        & \rstate_{k+1} = \mathbf{A}^j_k \rstate_k + \Br \ctrl_k + \mathbf{d}^j_k,  \forall k \in \{1, \dots, N\}, \nonumber \\
        & \perf_k = \mathbf{H}^j_k \rstate_k + \mathbf{c}^j_k, \quad\quad\quad\quad \forall k \in \{1, \dots, N\}, \\
        & \perf_k \in \Observable, \quad\quad\quad\quad\quad\quad\quad\,\, \forall k \in \{1, \dots, N\}, \nonumber \\
        & \ctrl_k \in \Cntrl, \quad\quad\quad\quad\quad\quad\quad\,\, \forall k \in \{1, \dots, N-1\}. \nonumber
\end{align}
The terms $\mathbf{A}^j_k \in \R^{\nrstate \times \nrstate}$ and $\mathbf{H}^j_k \in \R^{\nperf \times \nrstate}$ represent Jacobians of the dynamics and the observation map, respectively, while $\mathbf{d}^j_k \in \R^{\nrstate}$ and $\mathbf{c}^j_k \in \R^{\nperf}$ are the accompanying residuals defined as
\begin{align}
\label{eq:SCP}
& \mathbf{A}^j_k \defeq \frac{\partial{\mathbf{r}_{\text{d}}}}{\partial{\rstate}_k} \bigg\rvert_{\rstate_k=\rstate^j_k, \ctrl_k=\ctrl^j_k}, \quad \mathbf{d}_k \defeq \mathbf{r}_{\text{d}}(\rstate_k^j, \ctrl_k^j) - \mathbf{A}^j_k \rstate_k^j - \Br \ctrl^j_k, \nonumber \\
&\mathbf{H}^j_k \defeq \mathbf{C} \frac{\partial{\mathbf{w}}}{\partial{\rstate}_k} \bigg\rvert_{\rstate_k=\rstate^j_k, \ctrl_k=\ctrl^j_k}, \quad \mathbf{c}_k \defeq \mathbf{C} \mathbf{w}(\rstate_k^j, \ctrl_k^j) - \mathbf{C} \mathbf{H}^j_k \rstate^j_k,
\end{align}
where $\mathbf{r}_{\text{d}}: \R^\nrstate \rightarrow \R^\nrstate$ represents the time-discretized function of $\mathbf{r}$.
A sequence of \textbf{(LOCP)}$_{j}$ are solved until convergence \ie $\norm{\rstate^{j+1} - \rstate^j}_2 < \varepsilon$ for arbitrarily small $\varepsilon$. 

Solving \eqref{eq:SCP} requires each \textbf{(LOCP)}$_{j}$ to be feasible, which is always possible with the introduction of \textit{virtual dynamics}, as detailed in \cite{mao2016successive}.
Additionally, since linearization provides a good approximation to the nonlinear dynamics only in a small neighborhood around the nominal trajectory, we use \textit{trust regions} to ensure smooth convergence.
We implement a modified version of \cite{BonalliBylardEtAl2019} to solve the formulated SCP and treat the constraints on the performance variables as soft constraints.

We solve the finite-horizon problem in a receding horizon fashion, where each receding horizon subproblem involves solving the \textbf{(LOCP)}. This gives an optimal reduced-order model trajectory $(\bar{\rstate}^*, \ctrl^*) = (\{ \rstate^*_k\}_{k=1}^N, \{ \ctrl^*_k\}_{k=1}^{N-1})$ approximating the solution to OCP \eqref{eq:OCP} over an arbitrarily long, finite horizon.
}{
\newcounter{appendixsubsections}
\renewcommand{\theappendixsubsections}{\Alph{appendixsubsections}}
\refstepcounter{appendixsubsections}\label{appendix:2ndorder}
\refstepcounter{appendixsubsections}\label{appendix:SSM}
\refstepcounter{appendixsubsections}\label{appendix:SCP}
}
\end{document}